\documentclass[pdflatex,sn-mathphys-num]{sn-jnl}

\usepackage{graphicx}%
\usepackage{multirow}%
\usepackage{amsmath,amssymb,amsfonts}%
\usepackage{amsthm}%
\usepackage{mathrsfs}%
\usepackage[title]{appendix}%
\usepackage{xcolor}%
\usepackage{textcomp}%
\usepackage{manyfoot}%
\usepackage{booktabs}%
\usepackage{algorithm}%
\usepackage{algorithmicx}%
\usepackage{algpseudocode}%
\usepackage{listings}%
\usepackage{lineno}

\usepackage{algorithm}
\usepackage{algpseudocode}

\usepackage{afterpage}

\begin{document}

\title[Article Title]{SoftSnap: Rapid Prototyping of Untethered Soft Robots Using Snap-Together Modules}








\author*[1]{\fnm{Luyang} \sur{Zhao}}\email{luyang.zhao@dartmouth.edu}

\author[1]{\fnm{Yitao} \sur{Jiang}}\email{yitao.jiang@dartmouth.edu}

\author[1]{\fnm{Chun-Yi} \sur{She}}\email{chun-yi.she@dartmouth.edu}

\author[2]{\fnm{Muhao} \sur{Chen}}\email{muhao.chen@uky.edu}

\author*[1]{\fnm{Devin} \sur{Balkcom}}\email{devin.balkcom@dartmouth.edu}

\affil*[1]{\orgdiv{Department of Computer Science}, \orgname{Dartmouth College}, \orgaddress{\city{Hanover}, \state{NH}, \postcode{03755}, \country{USA}}}

\affil[2]{\orgdiv{Department of Mechanical and Aerospace Engineering}, \orgname{University of Kentucky}, \orgaddress{\city{Lexington}, \state{KY}, \postcode{40506}, \country{USA}}}


\abstract{

Soft robots offer adaptability and safe interaction with complex environments. 
Rapid prototyping kits that allow soft robots to be assembled easily will allow different geometries to be explored quickly to suit different environments or to mimic the motion of biological organisms.
We introduce \textit{SoftSnap} modules: snap-together components that enable the rapid assembly of a class of untethered soft robots. Each SoftSnap module includes embedded computation, motor-driven string actuation, and a flexible thermoplastic polyurethane (TPU) printed structure capable of deforming into various shapes based on the string configuration. These modules can be easily connected with other SoftSnap modules or customizable connectors.
We demonstrate the versatility of the SoftSnap system through four configurations: a starfish-like robot, a brittle star robot, a snake robot, a 3D gripper, and a ring-shaped robot. These configurations highlight the ease of assembly, adaptability, and functional diversity of the SoftSnap modules.
The SoftSnap modular system offers a scalable, snap-together approach to simplifying soft robot prototyping, making it easier for researchers to explore untethered soft robotic systems rapidly.

}

\maketitle
\section*{Introduction}
\label{sec:introduction}

\begin{figure}[htb]
    \centering
    \includegraphics[width=\linewidth]{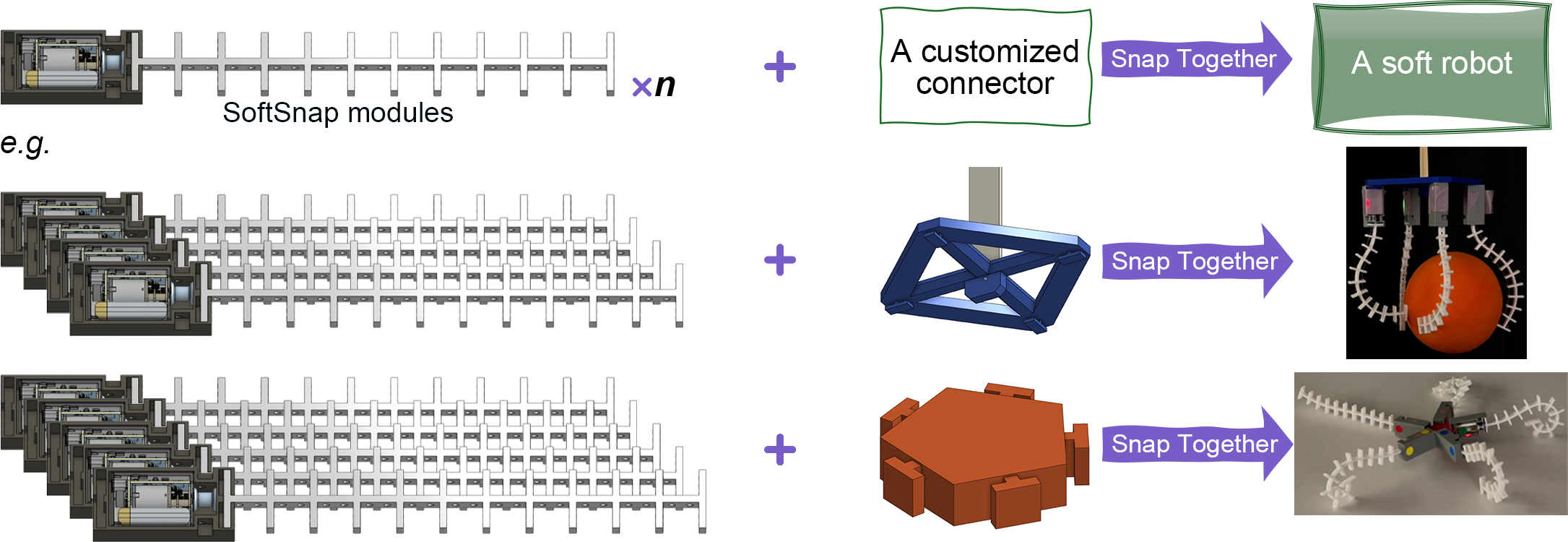}
    \caption{\textbf{Overview of the prototyping process.} The first row demonstrates that combining $n$ SoftSnap modules with a customized connector can construct a soft robot. The second row illustrates a gripper robot made from four SoftSnap modules arranged in a square and connected in the center. The third row depicts a brittle star-like robot created with five SoftSnap modules in a pentagonal configuration, also connected centrally.}
    \label{fig:overview}
\end{figure}

Soft robots are adaptable and compliant~\cite{rus2015design}, enabling the safe manipulation of delicate objects and navigation through challenging terrain, with applications in medical devices and search and rescue operations~\cite{dong2022recent}. 
Integrating controllers, batteries, actuators, and sensors into untethered, compliant systems poses a challenge due to the complexity of the assembly process~\cite{lee2017soft,Lipson2014ChallengesAO}. Current actuation technologies for soft robots include pneumatic~\cite{9720228,turtle}, cable-driven~\cite{chen2023analysis,9716747}, electrically-responsive~\cite{zebing1,lip2}, magnetically-responsive~\cite{mag}, and thermal-responsive systems (e.g., shape memory alloys)~\cite{zhao1,zhao2}. Pneumatic actuation, for example, is widely used for its simplicity and smooth, fluid-like motion, but it often requires external pumps and tubing, limiting its practicality in untethered applications~\cite{9785890,zhai,lip2,Tawk2021ARO}.
The selection of the actuation method informs the design of a rapid-prototyping kit.

Systems like Legos and similar kits have enabled rapid prototyping of articulated rigid-body structures or robots~\cite{8658751}. Modular robots, leveraging similar systems, can adapt their physical structure to different tasks through reconfiguration~\cite{liang2023decoding, Yim2002ModularRobots, reconfig_review, 4141032}. Most modular robots rely on rigid components~\cite{8769941, wei2010Sambot,multi_leg_swarm}. Recent research has begun integrating soft robotics principles into modular designs~\cite{freeman2023topology,Li2019JelloCubeAC,zou2018reconfigurable,ShapeBotsSuzukiEtAl}, aiming to develop reconfigurable soft robots that combine compliance and adaptability for diverse applications~\cite{Foambots,eciton,li2022scaling}. However, these approaches often involve intricate manual pre-assembly of modules, making it difficult to rapidly explore alternative geometries or to scale up to very large numbers of modules~\cite{ModularSR,2024_Lee,untethered_isoperimetric}. 
Simplifying the assembly process and designing soft modules that support rapid prototyping is, therefore, a promising area for research~\cite{schmitt2018soft,wehner2016integrated}.



\begin{figure*}[htbp]
    \centering
    \includegraphics[width=\linewidth]{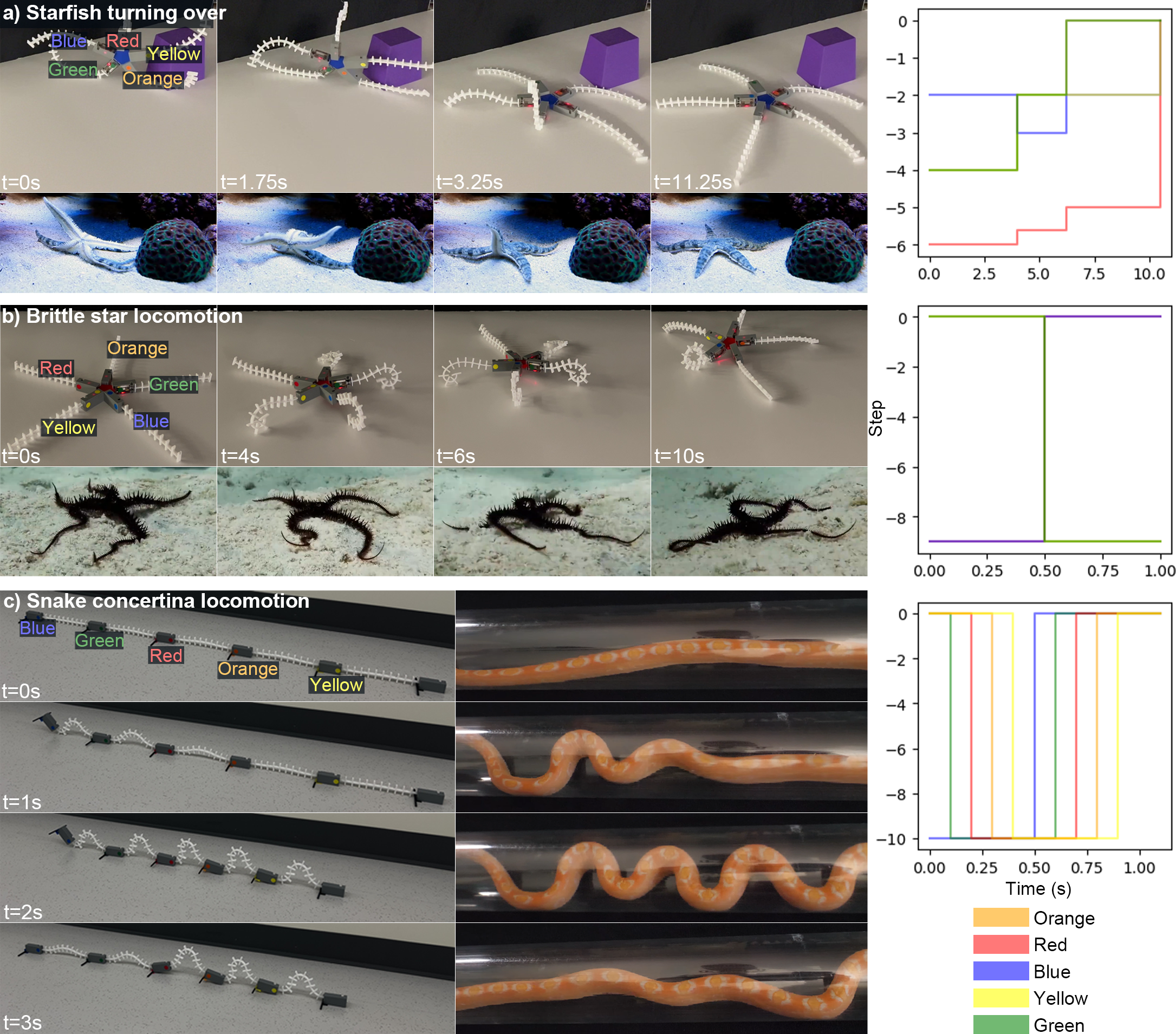}
    \caption{\textbf{Examples of the cable-driven SoftSnap modular system applied to bio-inspired robotic designs.} The image illustrates three robotic motions inspired by nature.
    The right images show control signals for five modules, highlighted in blue, red, yellow, orange, and green. \textbf{(a)} The starfish-like robot replicates the turning-over behavior observed in real starfish during 10.5 s. At t = 0 s, the robot leans on a purple block shaped to resemble a starfish. 
    By t = 1.75 s, the robot achieves a similar posture to a real starfish, with its top leg raised. It continues to receive control signals until it falls due to gravity at t = 3.25 s. The robot vividly emulates starfish motion until t = 11.25 s, after which the control is switched off, and the robot relaxes, resembling a starfish resting on the seafloor.
    \textbf{(b)} The brittle star-inspired robot demonstrates locomotion through periodic movements with a cycle of 1 s. Initially, the robot lies flat with its legs extended. A large step control signal is applied for 0.5 s, causing the legs to contract rapidly, followed by a second control signal for the remaining 0.5 s to release the actuation and contract the other legs. This stretch and shrink motion generates momentum, 
    showing clear displacement from its starting position.
    \textbf{(c)} The snake-like robot performs concertina locomotion by sequentially deforming its modules. In this demonstration, the robot exhibits a periodic motion with a cycle of 1.1 s. 
    It starts in a straight position, contracts its body, achieves full contraction, begins to unwind, and progressively returns to its initial straight state from head to tail. This is achieved by periodically alternating the step signals in different modules.
    The picture on the left shows that at t=0 s, the robot is in its initial state; at t=1 second, it begins to contract; at t=2 s, it is fully contracted; and by t=3 s, it is midway through returning to its original state. 
    Credits for the three real animal images: \textit{SeaAnimal4k} (Star fish 4k Amazing Starfish in Undersea Ultra Hd, \url{https://www.youtube.com/watch?v=MWlKfaxROd0}), \textit{keepondiving3140} (Black brittle star Small Giftun Red Sea Egypt, \url{https://www.youtube.com/watch?v=cX9c43sF3vo}), and \textit{SnakeDiscovery} (How Snakes Move! (They don't just slither!), \url{https://www.youtube.com/watch?v=7-AKPFiIEEw&t=30s}).}
    \label{fig:applications1}
\end{figure*}

\begin{figure*}[htbp]
    \centering
    \includegraphics[width=\linewidth]{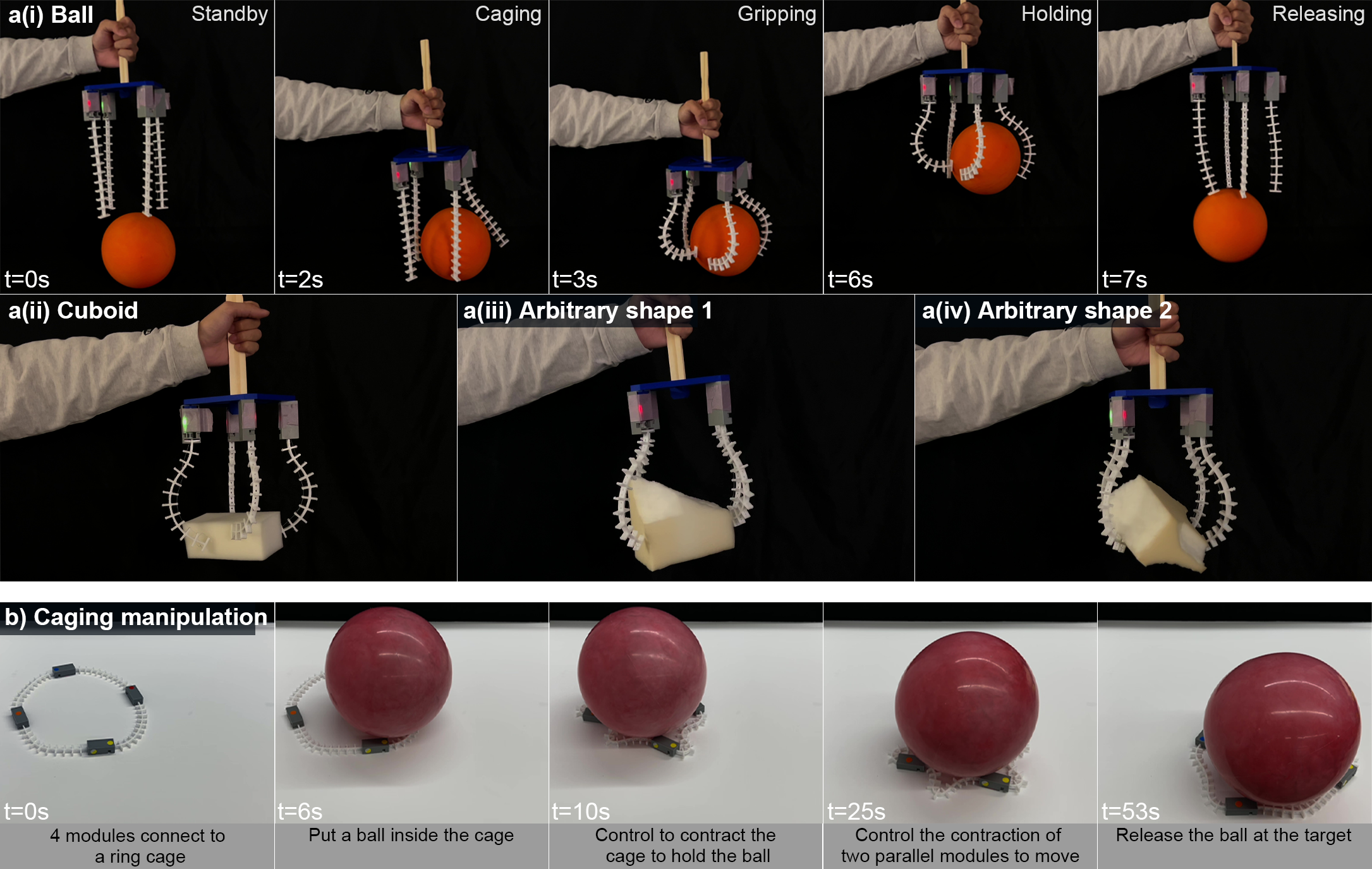}
    \caption{\textbf{Demonstrations of the cable-driven modular system applied to manipulation tasks.} \textbf{(a)} The gripper robot effectively grasps objects of different shapes, such as a ball and a cuboid. In \textbf{a(i)}, we illustrate the five stages of ball grasping. In \textbf{a(ii)}, \textbf{a(iii)}, and \textbf{a(iv)}, we demonstrate grasping with three body shapes: one cuboid and two irregular forms, all controlled by the same signal for the four modules. \textbf{(b)} A four-module ring configuration is used for caging-type object transport, showcasing the ability to transport a large ball. At t = 0 s, the ring cage is in its initial state.}
    \label{fig:applications2}
\end{figure*}

This paper introduces a technique and components for quickly prototyping a certain class of soft robots with different geometries.
The system is primarily intended for scientists with a range of backgrounds interested in exploring robotic analogs of biological organisms, or soft robot geometries suitable for manipulation or locomotion tasks. The system enables easy component reuse and rapid testing of different ideas about mechanical design, control, and applications.
We detail the design and initial implementation of cable-driven, modular soft actuator units 
that can be snapped together to quickly prototype untethered soft robotic systems. There are two main components: (1) actuatable SoftSnap modules that each integrate a battery, motor, Wi-Fi communication, and control system, and (2) connectors that snap the modules together. Connectors can be easily 3D-printed to meet specific application requirements and can be re-used once built. An overview of the system is given in Fig.~\ref{fig:overview}, and Fig.~\ref{fig:applications1} shows some biologically-inspired example robots we constructed with the system. The SoftSnap modules, in some ways, serve a similar purpose to direct-driven joints in a traditional robot arm: the modules have one degree of freedom actuated by changing the cable length. Higher-degree-of-freedom serial or parallel structures for manipulation or locomotion can be formed by connecting these modules in various topologies using the connectors.

Each SoftSnap module incorporates a passive skeleton with holes through which the actuating cable is threaded. Using different holes leads to different {\em threading configurations}, each leading to a different deformation pattern for the module. Snapping modules together is trivially easy; the slowest steps of building a new design include threading modules and possibly printing custom connectors. Once modules and connectors of sufficient variety have been constructed, they form a kit that allows very fast exploration of new soft robot structures.

Our current platform is meant to enable human designers to imagine, duplicate, or explore soft robot geometries. One difficulty in the process is that human designers may not have good intuition about what shapes a module with a particular threading pattern might achieve during actuation. We, therefore, present a quasi-static simulator that minimizes potential energy to predict the shapes achieved for different string lengths for a particular threading of the module. 
We also modified the simulator for use as a design tool. This tool allows users to input a desired curve, and then attempts to find the threading design that most closely matches the curve.

We demonstrate the versatility of the snap-together soft modular system through four distinct example assemblies: a gripper, a starfish robot, a snake robot, and a ring-shaped robot. These configurations showcase how the snap-together modules can be easily reassembled to create a range of functional soft robots, enabling untethered operation and supporting rapid prototyping, as shown in Figs.~\ref{fig:applications1} and \ref{fig:applications2}. In this paper, we detail the design, modeling, and experimental validation of these modular soft robots, highlighting their adaptability for diverse applications.

\section*{Results}

\begin{figure*}[htbp]
    \center
    \includegraphics[width=\linewidth]{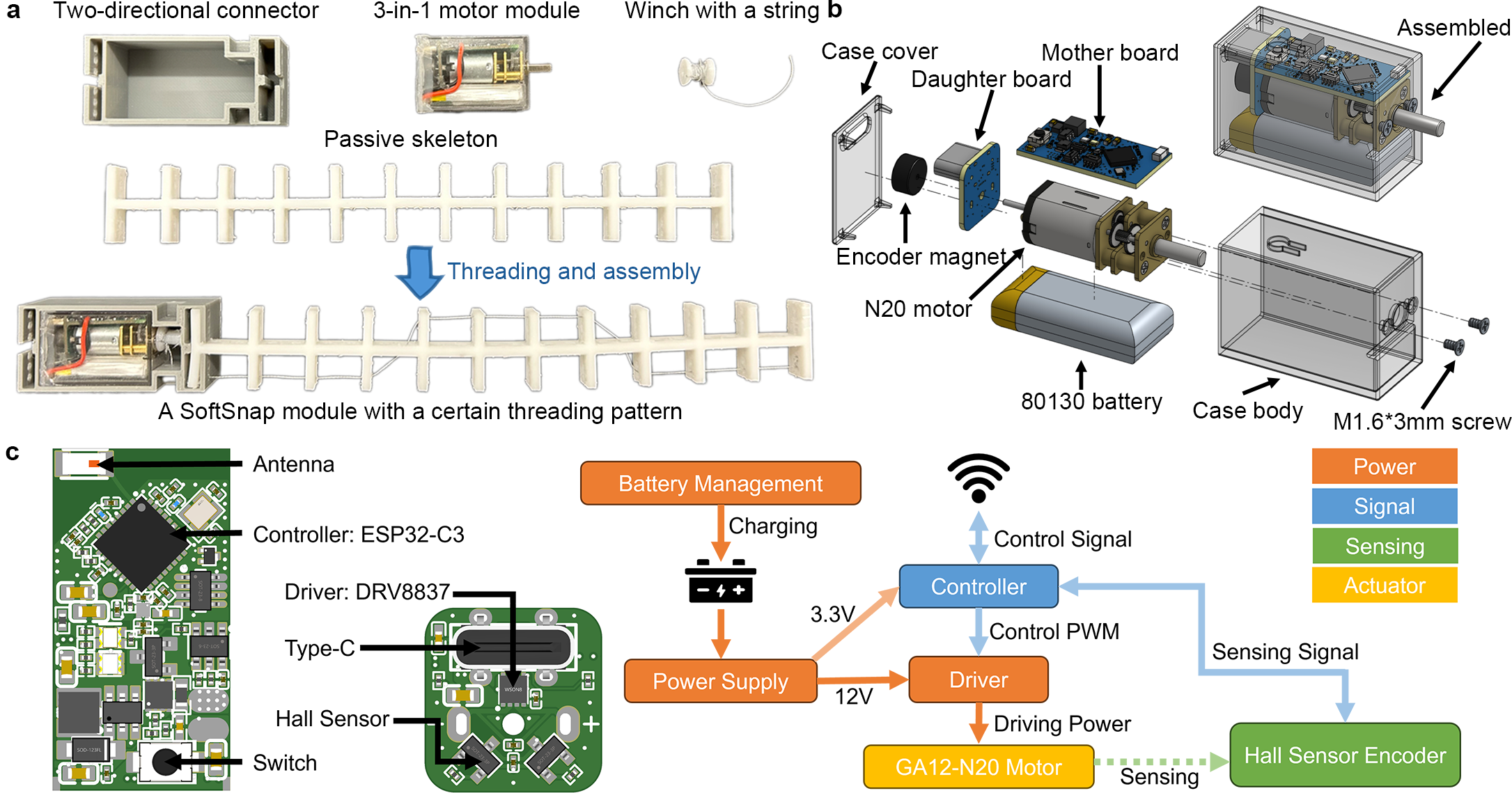} %
    \caption{\textbf{SoftSnap module design and motor module details.} \textbf{(a)} Three components of the actuatable module: passive skeleton, 3-in-1 motor module, and two-directional connector. The passive skeleton is linked to the 3-in-1 motor module via a winch. A string extends from the motor, passes over the winch, threads through the holes on the passive skeleton, and continues to the far end of the passive skeleton. \textbf{(b)} The exploded-view drawing of the 3-in-1 motor module components which satisfied the three functionalities: Power, Control, and Actuation.
    \textbf{(c)} The PCB's mother board and daughter board are displayed on the left. The system function flow is present on the right.}
    \label{fig:combined-figure}
\end{figure*}



\subsection*{Module and connector design}

Each SoftSnap module houses all electronic components in a compact box connected to a passive skeleton. Each spine of the skeleton has holes that can be used to thread the single cable; different selections of holes lead to different threading patterns and different deformations when the cable is shortened. As shown in Fig.~\ref{fig:combined-figure}(a), the module includes: (i) {Passive Skeleton} with a uniformly designed structure featuring adjustable rib lengths, thickness, and middle rod dimensions for customized mechanical properties. (ii) {3-in-1 Motor Module} (Fig.~\ref{fig:combined-figure}(b)) combines a motor, PCB (Fig.~\ref{fig:combined-figure}(c)), and battery in one compact unit for untethered operation and Wi-Fi control, including USB Type-C charging. (iii) {Two-Directional Connector} connects the passive skeleton to the motor module and allows for easy modular assembly via a snap-together mechanism. For example, as shown in Figs.~\ref{fig:applications1} and \ref{fig:applications2}, it supports the assembly of modules in configurations such as a chain (a snake-like robot) or a ring-shaped robot.

Multiple SoftSnap modules can combine with customized connectors to form various soft robots, as shown in Fig.~\ref{fig:overview}.
These connectors facilitate various configurations by linking different numbers of modules in diverse positions and orientations. For instance, as shown in Fig.~\ref{fig:applications1}(a), the starfish configuration employs a 3D pentagonal connector with small connectors to arrange five modules in different orientations. In the gripper configuration, shown in Fig.~\ref{fig:applications2}(a), a cuboid connector arranges four modules symmetrically.

\begin{figure*}[htbp]
    \center
    \includegraphics[width=0.95\linewidth]{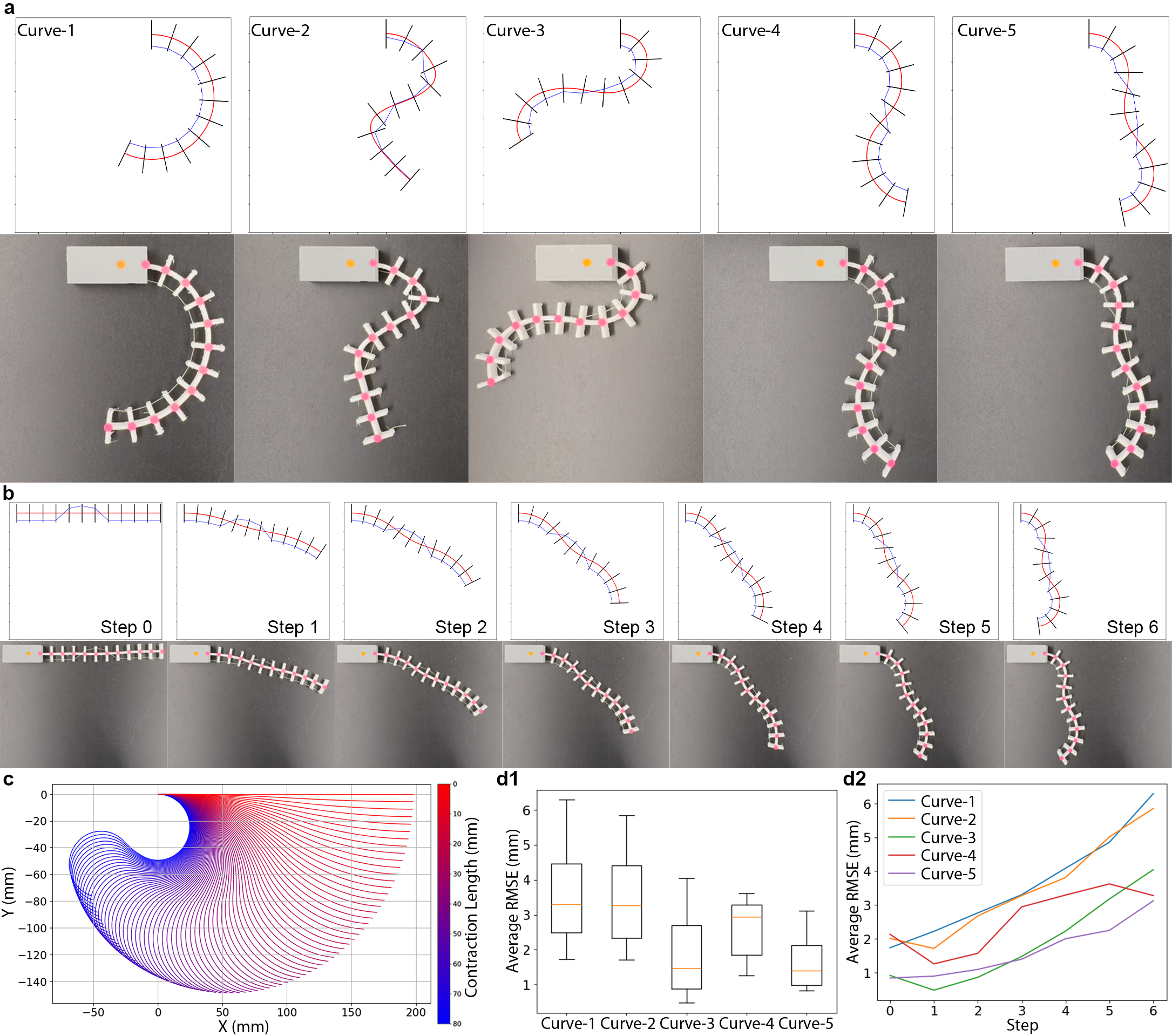} 
    \caption{\textbf{Five different threading methods and their resulting deformations after shortening the strings to the same length in both simulations and real experiments. }\textbf{(a)} The left end of the skeleton, attached to the body case, is fixed to the ground. The blue line illustrates the single string that runs from the motor to the skeleton and reaches its far end. The red line represents the deformed shape of the skeleton along its longitudinal axis. \textbf{(b)} Example of a \lq\lq w" (curve-5) shape configuration, comparing seven steps of shortening the strings evenly in experiments and simulations. \textbf{(c)} Displays all the configurations with the `Curve-3' threading method and shows the workspace of the middle of the skeleton from 0 mm to 80 mm string contract length; the contraction process is associated with the color bar. \textbf{(d1)} and \textbf{(d2)} The average RMSE is calculated with the RMSE of the 12 center points (marked in pink) of each rod on the skeleton. 
    \textbf{(d1)} Illustrates the average RMSE across five experiment threading patterns across seven testing steps. Within the experiment range, the real robot shows a good match with the simulation. \textbf{(d2)} Shows the RMSE of five different experiment threading patterns over steps. The average RMSE, with a maximum of about 6 mm, is relatively small compared to the skeleton's length of 201 mm. The error has an increment trend with the string length contraction.}
    \label{fig:five-skeleton}
\end{figure*}

\subsection*{Quasi-static modeling and simulation}
\label{sec:modeling}

To understand how different threading patterns influence the deformation and configuration space, we developed a quasi-static model that approximates the deformation behavior for a given threading pattern.

\textbf{Forward quasi-static modeling}: Given a threading pattern \( T \) and target string length \( L_{\text{target}} \), we compute the resulting deformation shape of the skeleton.  We can also explore the complete set of deformations for a module by iteratively reducing the string length, using the geometry from the previous iteration as the initial guess for the next step.

The skeleton is modeled as a collection of segments, with each pair of adjacent ribs storing potential energy to resist bending. At each iteration, a trust-region-constrained optimizer
minimizes the total potential energy, subject to the string length constraint. The potential energy is modeled as the sum of squared bending angles \( \alpha_i \) between adjacent ribs. Given a threading pattern \( T_i \) for the \( i \)-th pair of ribs, the string length \( L_i \) is calculated as a function of the bending angle \( \alpha_i \), and the target total string length \( L_{\text{target}} \) must satisfy \( L_{\text{target}} < L_{\text{original}} \). The optimization problem is formulated as:
\begin{align}
    \min \sum_{i} \alpha_i^2 \quad \text{s.t.} \quad \sum_{i} L_i = \sum_{i} L(\alpha_i, T_i) = L_{\text{target}}.
\end{align}

Although \( L_i \) can be directly calculated from \( \alpha_i \), the reverse is more complex due to the non-linear relationship. To find \( \alpha_i \) given \( L_{\text{target}} \), numerical methods like Newton-Raphson
can be employed. The relationship between \( L_i \) and \( \alpha_i \) is detailed in the Supplementary ``String Length Calculation" section.

Thus, the optimization procedure iteratively updates \( \alpha_i \), ensuring the total string length matches \( L_{\text{target}} \) while minimizing potential energy to find the optimal bending configuration. The complete procedure is outlined in Algorithm 1.



\vspace{1em} 
\noindent\textbf{Inverse quasi-static modeling}: Given a target deformation shape, a human designer may wish to determine the threading pattern and corresponding string length. We simplify the input by representing the desired deformation as a sequence of target bending angles (\( \alpha_0 \), \( \alpha_1 \), \( \cdots \), \( \alpha_n \)). Using these angles, the positions of all intersection points between the ribs and the middle rods are explicitly calculated based on geometric relations. Suppose each rib has \( m \) holes, labeled as \( t_1 \), \( t_2 \), \( \cdots \), \( t_m \). For each pair of adjacent ribs (each segment), there are \( m^2 \) possible threading patterns. For a specific combination of threading, such as \( t_i \) and \( t_j \), we calculate the corresponding string length \( L_{ij} \) based on the required bending angle. In addition to the threading pattern, we also store the string length for each segment. If \( L_{ij} \) exceeds the original string length (the length between the holes when the angle is zero), that threading pattern is discarded. This process yields a set of valid threading options for each pair of ribs.

Next, we applied a backtracking method to find all valid threading paths from the first rib to the final rib. This method ensures that each threading path is consistent with the geometrical constraints of adjacent ribs. During backtracking, we also calculate the total string length for each threading path by summing the string lengths of all segments. Once all valid paths are generated, we iterate through them, using the threading pattern and the total string length of each path as inputs. For each path, we apply the forward quasi-static model to compute the resulting bending angles. We then calculate the distance between these resulting angles (in radians) and the target angles using the L2 norm. The path that produces the minimum distance is selected as the optimal solution. The algorithm outputs the corresponding threading pattern and the total string length of the chosen path.

There are several key points to note. First, the number of holes on each rib significantly affects computational cost; for example, with 3 holes per rib, the total number of possible threading patterns for \( n \) ribs is \( 3^n \). Although the filtering method reduces the number of valid threadings, multiple configurations may still achieve the same target angle, leaving many possible solutions. To address this in practice, parallel computing can be employed, or a maximum number of threading options can be set, with the best solution chosen from this subset. Figure S4 (a) and (b) demonstrate that different threading patterns and string lengths can produce the same desired angle sequence. Second, some angles may be unachievable due to physical constraints of the threading hole positions, and the inverse algorithm finds the closest solution given the number of holes on each rib, though it may not always perfectly match the desired angles. Figures S4 (c) and (d) illustrate examples where 8 holes per rib were used, and the best solution was found within these threading constraints.

\vspace{1em} 
\noindent\textbf{Comparison between simulation and experiments}:
To validate the skeleton design and compare the robot's physical behavior with simulations, we employed five different threading methods, as shown in Fig.~\ref{fig:five-skeleton}. Color-coded markers (pink and orange) were placed at the intersections of ribs and middle rods. This tracked the skeleton's deformation for comparison with simulated results. We captured top-down images of the skeleton using a calibrated camera to ensure accurate spatial measurements. Standard computer vision techniques, such as color thresholding to isolate the pink and orange markers, contour detection to identify marker boundaries, and image moments to calculate marker centroids, were employed. These centroids provided the $x$ and $y$ coordinates of the markers. The line connecting the pink and orange markers defined the $x$-axis, and the image was rotated to align the skeleton with the horizontal axis for consistent analysis.

After aligning the skeleton, we measured the adjusted marker positions relative to the first pink marker, which is closest to the orange marker and is designated as the origin. These measurements allowed us to evaluate how various threading methods influenced the skeleton’s deformation. The experimental results were then compared with the simulation data to assess the model's accuracy and examine how threading variations and string lengths affected the robot’s performance. Moreover, we also illustrate the five different threading methods and the resulting deformations when the strings are shortened to equal lengths, both in simulation and real experiments. 

\subsection*{Case studies and experimental validation}
\label{sec:applications}

In this section, we present case studies and experimental results. 


\subsubsection*{Starfish and brittle star robots}

To create a starfish-like robot, we designed a 3D pentagonal connector with small pegs for securely attaching the connectors of the actuatable modules. Two connector configurations were developed: one (red connector) positions all modules with their deformation direction perpendicular to the connector, while the other (blue connector) aligns some modules perpendicular and others parallel to the connector.
Fig.~\ref{fig:applications1}(a) and Movie S1 illustrate how the robot mimics the righting response of a real starfish by deforming its five limbs. In contrast, Fig.~\ref{fig:applications1}(b) and Movie S1 demonstrate locomotion that closely resembles the movement of brittle stars, achieved by coordinating the deformation of its limbs in sequences. The control sequence for each skeleton is illustrated in Fig.~\ref{fig:applications1}. Both robots use the same threading patterns for the five skeletons, with details of the threading pattern provided in Figure S3.

\subsubsection*{Snake-like robot}

The snake-like robot was constructed by connecting a chain of five skeleton modules, each threaded in a consistent pattern to form an ``$\omega$" shape (Figure S3). To enhance ground stability, six short rods were attached to the connectors.  Figure~\ref{fig:applications1}(c) and Movie S1 showed an example. Forward movement was achieved by sequentially deforming the modules from front to back, then restoring them to their original shape, mimicking the concertina locomotion observed in real snakes, especially in confined spaces or rough terrain.

\subsubsection*{Gripper}

The gripper was developed by attaching four actuatable modules to a rectangular connector arranged in an ``X" configuration, as shown in Figure~\ref{fig:applications2}(a) and Movie S1. By adjusting the connector design and threading pattern, different types of grippers can be produced. In this example, the modules were threaded to form an ``S" shape (Figure S3), with a larger lower curve and a smaller upper curve. This setup allows the gripper to grasp objects of various shapes, such as balls, cuboids, and irregular objects. However, the current design is only suitable for handling lightweight items. Handling heavier objects would require a stiffer skeleton or perhaps a different threading configuration.

\subsubsection*{Caging-type object transport}

A caging-type object transport system was developed by connecting four modules into a ring configuration, as shown in Figure~\ref{fig:applications2}(b) and Movie S1. The threading pattern used is the same as that for the snake-like robot. This system was used to transport a lightweight ball, using sequential deformation of the modules. Initially, two modules were actuated to deform and grip the ball, after which the remaining two modules returned to their original shape and were then deformed again to continue the transport. This alternating deformation allowed for continuous object movement.
While the current system successfully demonstrated the transport of lightweight objects, such as the tested ball, transporting heavier objects would require further modification, such as increasing the stiffness of the skeleton.

\section*{Discussion}

Soft robotics systems often encounter challenges in rapid prototyping and achieving untethered, actuatable modules adaptable to various applications. Our research aims to address this by introducing a cable-driven modular system that facilitates the quick and scalable assembly of soft robots using snap-together components. We have integrated key electronic components—such as communication hardware, a battery, and a motor—into a compact box that combines with a flexible skeleton to form an untethered, actuated, and bendable module. These modules can be easily snapped together to create different soft robotic configurations, such as starfish-like, snake-like, and gripper robots.
This approach helps to reduce the complexity of soft robot manufacturing, allowing researchers to focus more on design exploration and application functionality.

A key insight from our research is the role of threading patterns in shaping the skeleton’s deformation. We developed a model demonstrating how different threading configurations affect the overall shape and flexibility of the skeleton. By simply adjusting the threading, users can achieve various deformations, which can then be used to prototype distinct robot forms. This capability allows researchers to focus more on control strategies and specific applications rather than the intricacies of the construction process.

Despite these advancements, several limitations remain. The current work focused primarily on the effects of threading on deformation patterns, while other critical factors—such as material properties of the skeleton, rib dimensions, and mechanical load capacities—were not fully explored. Additionally, the skeleton design operates mainly in 2D, limiting the complexity and range of motions. Expanding the system into a full 3D configuration and incorporating multiple strings per module could provide greater control and precision, broadening the scope of possible applications. Moreover, advanced control algorithms, including real-time feedback mechanisms and adaptive behaviors, should be explored to improve the system's responsiveness to environmental changes.

Our research provides a starting point for a more accessible and scalable method of soft robot development. By simplifying the construction process, researchers can now focus on exploring new forms, behaviors, and control strategies for their robots. We anticipate that SoftSnap modules will enable rapid prototyping for a range of applications, from biomimetic locomotion to manipulation tasks, while also lowering the barrier to entry for soft robotic research. Future work will focus on enhancing the design with multi-string actuation for finer control, as well as conducting field tests to assess the system's robustness and scalability in real-world environments.

\section*{Methods}

\subsection*{SnapSoft module and connector fabrication}

A SnapSoft module is composed of three main components. 1) The passive skeleton is 3D-printed using TPU filament (1.75 mm, Amazon Basics), selected for its flexibility, durability, and ability to deform under applied loads.
2) The 3-in-1 motor module integrates a motor, a custom-designed printed circuit board (PCB), and an 80130 lithium-ion battery (3.7 V, 200 mAh, 30C). The battery’s high energy density allows for satisfactory operation while minimizing bulk. The PCB, as shown in Figure~\ref{fig:combined-figure}(c), incorporates the controller and driver and soldered two parts perpendicular to minimize bulk. The motor, type GA12-N20 (50,000 RPM) with a gear ratio of 1:100, was selected for its balance between torque and speed. All of these components are housed in a compact casing, 3D-printed using resin material, ensuring efficient integration with the skeleton.
3) The two-directional connectors, which are 3D-printed using PLA filament (1.75 mm, Overture), enable modular reconfiguration of the skeleton. These connectors allow for the formation of different structures, depending on the application. The snap-together mechanism has been tested for durability, ensuring reliable connections between modules and the connector structure.

The customized connectors are fabricated from polylactic acid (PLA) filament (1.75 mm, Overture), chosen for its rigidity and compatibility with other components.


\section*{Data availability} 

All data needed to evaluate the conclusions in the paper are present in the paper or the Supplementary Materials. Computer code is available from the first author and corresponding authors upon request.

\bibliography{references}


\begin{thebibliography}{36}
\ifx \bisbn   \undefined \def \bisbn  #1{ISBN #1}\fi
\ifx \binits  \undefined \def \binits#1{#1}\fi
\ifx \bauthor  \undefined \def \bauthor#1{#1}\fi
\ifx \batitle  \undefined \def \batitle#1{#1}\fi
\ifx \bjtitle  \undefined \def \bjtitle#1{#1}\fi
\ifx \bvolume  \undefined \def \bvolume#1{\textbf{#1}}\fi
\ifx \byear  \undefined \def \byear#1{#1}\fi
\ifx \bissue  \undefined \def \bissue#1{#1}\fi
\ifx \bfpage  \undefined \def \bfpage#1{#1}\fi
\ifx \blpage  \undefined \def \blpage #1{#1}\fi
\ifx \burl  \undefined \def \burl#1{\textsf{#1}}\fi
\ifx \doiurl  \undefined \def \doiurl#1{\url{https://doi.org/#1}}\fi
\ifx \betal  \undefined \def \betal{\textit{et al.}}\fi
\ifx \binstitute  \undefined \def \binstitute#1{#1}\fi
\ifx \binstitutionaled  \undefined \def \binstitutionaled#1{#1}\fi
\ifx \bctitle  \undefined \def \bctitle#1{#1}\fi
\ifx \beditor  \undefined \def \beditor#1{#1}\fi
\ifx \bpublisher  \undefined \def \bpublisher#1{#1}\fi
\ifx \bbtitle  \undefined \def \bbtitle#1{#1}\fi
\ifx \bedition  \undefined \def \bedition#1{#1}\fi
\ifx \bseriesno  \undefined \def \bseriesno#1{#1}\fi
\ifx \blocation  \undefined \def \blocation#1{#1}\fi
\ifx \bsertitle  \undefined \def \bsertitle#1{#1}\fi
\ifx \bsnm \undefined \def \bsnm#1{#1}\fi
\ifx \bsuffix \undefined \def \bsuffix#1{#1}\fi
\ifx \bparticle \undefined \def \bparticle#1{#1}\fi
\ifx \barticle \undefined \def \barticle#1{#1}\fi
\bibcommenthead
\ifx \bconfdate \undefined \def \bconfdate #1{#1}\fi
\ifx \botherref \undefined \def \botherref #1{#1}\fi
\ifx \url \undefined \def \url#1{\textsf{#1}}\fi
\ifx \bchapter \undefined \def \bchapter#1{#1}\fi
\ifx \bbook \undefined \def \bbook#1{#1}\fi
\ifx \bcomment \undefined \def \bcomment#1{#1}\fi
\ifx \oauthor \undefined \def \oauthor#1{#1}\fi
\ifx \citeauthoryear \undefined \def \citeauthoryear#1{#1}\fi
\ifx \endbibitem  \undefined \def \endbibitem {}\fi
\ifx \bconflocation  \undefined \def \bconflocation#1{#1}\fi
\ifx \arxivurl  \undefined \def \arxivurl#1{\textsf{#1}}\fi
\csname PreBibitemsHook\endcsname

\bibitem[\protect\citeauthoryear{Rus and Tolley}{2015}]{rus2015design}
\begin{barticle}
\bauthor{\bsnm{Rus}, \binits{D.}},
\bauthor{\bsnm{Tolley}, \binits{M.}}:
\batitle{Design, fabrication and control of soft robots}.
\bjtitle{Nature}
\bvolume{521},
\bfpage{467}--\blpage{75}
(\byear{2015})
\doiurl{10.1038/nature14543}
\end{barticle}
\endbibitem

\bibitem[\protect\citeauthoryear{Dong et~al.}{2022}]{dong2022recent}
\begin{barticle}
\bauthor{\bsnm{Dong}, \binits{X.}},
\bauthor{\bsnm{Luo}, \binits{X.}},
\bauthor{\bsnm{Zhao}, \binits{H.}},
\bauthor{\bsnm{Qiao}, \binits{C.}},
\bauthor{\bsnm{Li}, \binits{J.}},
\bauthor{\bsnm{Yi}, \binits{J.}},
\bauthor{\bsnm{Yang}, \binits{L.}},
\bauthor{\bsnm{Oropeza}, \binits{F.J.}},
\bauthor{\bsnm{Hu}, \binits{T.S.}},
\bauthor{\bsnm{Xu}, \binits{Q.}},
\bauthor{\bsnm{Zeng}, \binits{H.}}:
\batitle{Recent advances in biomimetic soft robotics: fabrication approaches{,} driven strategies and applications}.
\bjtitle{Soft Matter}
\bvolume{18},
\bfpage{7699}--\blpage{7734}
(\byear{2022})
\doiurl{10.1039/D2SM01067D}
\end{barticle}
\endbibitem

\bibitem[\protect\citeauthoryear{Lee et~al.}{2017}]{lee2017soft}
\begin{barticle}
\bauthor{\bsnm{Lee}, \binits{C.}},
\bauthor{\bsnm{Kim}, \binits{M.}},
\bauthor{\bsnm{Kim}, \binits{Y.J.}},
\bauthor{\bsnm{Hong}, \binits{N.}},
\bauthor{\bsnm{Ryu}, \binits{S.}},
\bauthor{\bsnm{Kim}, \binits{H.J.}},
\bauthor{\bsnm{Kim}, \binits{S.}}:
\batitle{Soft robot review}.
\bjtitle{International Journal of Control, Automation and Systems}
\bvolume{15}(\bissue{1}),
\bfpage{3}--\blpage{15}
(\byear{2017})
\doiurl{10.1007/s12555-016-0462-3}
\end{barticle}
\endbibitem

\bibitem[\protect\citeauthoryear{Lipson}{2014}]{Lipson2014ChallengesAO}
\begin{barticle}
\bauthor{\bsnm{Lipson}, \binits{H.}}:
\batitle{Challenges and opportunities for design, simulation, and fabrication of soft robots}.
\bjtitle{Soft Robotics}
\bvolume{1},
\bfpage{21}--\blpage{27}
(\byear{2014})
\doiurl{10.1089/soro.2013.0007}
\end{barticle}
\endbibitem

\bibitem[\protect\citeauthoryear{Kobayashi et~al.}{2022}]{9720228}
\begin{barticle}
\bauthor{\bsnm{Kobayashi}, \binits{R.}},
\bauthor{\bsnm{Nabae}, \binits{H.}},
\bauthor{\bsnm{Endo}, \binits{G.}},
\bauthor{\bsnm{Suzumori}, \binits{K.}}:
\batitle{Soft tensegrity robot driven by thin artificial muscles for the exploration of unknown spatial configurations}.
\bjtitle{IEEE Robotics and Automation Letters}
\bvolume{7}(\bissue{2}),
\bfpage{5349}--\blpage{5356}
(\byear{2022})
\doiurl{10.1109/LRA.2022.3153700}
\end{barticle}
\endbibitem

\bibitem[\protect\citeauthoryear{Baines et~al.}{2022}]{turtle}
\begin{barticle}
\bauthor{\bsnm{Baines}, \binits{R.}},
\bauthor{\bsnm{Patiballa}, \binits{S.}},
\bauthor{\bsnm{Booth}, \binits{J.}},
\bauthor{\bsnm{Ramirez}, \binits{L.}},
\bauthor{\bsnm{Sipple}, \binits{T.}},
\bauthor{\bsnm{Garcia}, \binits{A.}},
\bauthor{\bsnm{Fish}, \binits{F.}},
\bauthor{\bsnm{Kramer-Bottiglio}, \binits{R.}}:
\batitle{Multi-environment robotic transitions through adaptive morphogenesis}.
\bjtitle{Nature}
\bvolume{610},
\bfpage{283}--\blpage{289}
(\byear{2022})
\doiurl{10.1038/s41586-022-05188-w}
\end{barticle}
\endbibitem

\bibitem[\protect\citeauthoryear{Chen et~al.}{2023}]{chen2023analysis}
\begin{barticle}
\bauthor{\bsnm{Chen}, \binits{M.}},
\bauthor{\bsnm{Fraddosio}, \binits{A.}},
\bauthor{\bsnm{Micheletti}, \binits{A.}},
\bauthor{\bsnm{Pavone}, \binits{G.}},
\bauthor{\bsnm{Piccioni}, \binits{M.D.}},
\bauthor{\bsnm{Skelton}, \binits{R.E.}}:
\batitle{Analysis of clustered cable-actuation strategies of v-expander tensegrity structures}.
\bjtitle{Engineering Structures}
\bvolume{296},
\bfpage{116868}
(\byear{2023})
\doiurl{10.1016/j.engstruct.2023.116868}
\end{barticle}
\endbibitem

\bibitem[\protect\citeauthoryear{Lai et~al.}{2022}]{9716747}
\begin{barticle}
\bauthor{\bsnm{Lai}, \binits{J.}},
\bauthor{\bsnm{Lu}, \binits{B.}},
\bauthor{\bsnm{Zhao}, \binits{Q.}},
\bauthor{\bsnm{Chu}, \binits{H.K.}}:
\batitle{Constrained motion planning of a cable-driven soft robot with compressible curvature modeling}.
\bjtitle{IEEE Robotics and Automation Letters}
\bvolume{7}(\bissue{2}),
\bfpage{4813}--\blpage{4820}
(\byear{2022})
\doiurl{10.1109/LRA.2022.3152318}
\end{barticle}
\endbibitem

\bibitem[\protect\citeauthoryear{Mao et~al.}{2023}]{zebing1}
\begin{barticle}
\bauthor{\bsnm{Mao}, \binits{Z.}},
\bauthor{\bsnm{Peng}, \binits{Y.}},
\bauthor{\bsnm{Hu}, \binits{C.}},
\bauthor{\bsnm{Ding}, \binits{R.}},
\bauthor{\bsnm{Yamada}, \binits{Y.}},
\bauthor{\bsnm{Maeda}, \binits{S.}}:
\batitle{Soft computing-based predictive modeling of flexible electrohydrodynamic pumps}.
\bjtitle{Biomimetic Intelligence and Robotics}
\bvolume{3}(\bissue{3}),
\bfpage{100114}
(\byear{2023})
\doiurl{10.1016/j.birob.2023.100114}
\end{barticle}
\endbibitem

\bibitem[\protect\citeauthoryear{Miriyev et~al.}{2017}]{lip2}
\begin{botherref}
\oauthor{\bsnm{Miriyev}, \binits{A.}},
\oauthor{\bsnm{Stack}, \binits{K.}},
\oauthor{\bsnm{Lipson}, \binits{H.}}:
Soft material for soft actuators.
Nature Communications
\textbf{8}
(2017)
\doiurl{10.1038/s41467-017-00685-3}
\end{botherref}
\endbibitem

\bibitem[\protect\citeauthoryear{Chung et~al.}{2021}]{mag}
\begin{barticle}
\bauthor{\bsnm{Chung}, \binits{H.-J.}},
\bauthor{\bsnm{Parsons}, \binits{A.M.}},
\bauthor{\bsnm{Zheng}, \binits{L.}}:
\batitle{Magnetically controlled soft robotics utilizing elastomers and gels in actuation: A review}.
\bjtitle{Advanced Intelligent Systems}
\bvolume{3}(\bissue{3}),
\bfpage{2000186}
(\byear{2021})
\doiurl{10.1002/aisy.202000186}
{\href{https://arxiv.org/abs/https://onlinelibrary.wiley.com/doi/pdf/10.1002/aisy.202000186}{{https://onlinelibrary.wiley.com/doi/pdf/10.1002/aisy.202000186}}}
\end{barticle}
\endbibitem

\bibitem[\protect\citeauthoryear{Zhao et~al.}{2023}]{zhao1}
\begin{barticle}
\bauthor{\bsnm{Zhao}, \binits{L.}},
\bauthor{\bsnm{Wu}, \binits{Y.}},
\bauthor{\bsnm{Yan}, \binits{W.}},
\bauthor{\bsnm{Zhan}, \binits{W.}},
\bauthor{\bsnm{Huang}, \binits{X.}},
\bauthor{\bsnm{Booth}, \binits{J.}},
\bauthor{\bsnm{Mehta}, \binits{A.}},
\bauthor{\bsnm{Bekris}, \binits{K.}},
\bauthor{\bsnm{Kramer-Bottiglio}, \binits{R.}},
\bauthor{\bsnm{Balkcom}, \binits{D.}}:
\batitle{Starblocks: Soft actuated self-connecting blocks for building deformable lattice structures}.
\bjtitle{IEEE Robotics and Automation Letters}
\bvolume{8}(\bissue{8}),
\bfpage{4521}--\blpage{4528}
(\byear{2023})
\doiurl{10.1109/LRA.2023.3284361}
\end{barticle}
\endbibitem

\bibitem[\protect\citeauthoryear{Zhao et~al.}{2022}]{zhao2}
\begin{barticle}
\bauthor{\bsnm{Zhao}, \binits{L.}},
\bauthor{\bsnm{Wu}, \binits{Y.}},
\bauthor{\bsnm{Blanchet}, \binits{J.}},
\bauthor{\bsnm{Perroni-Scharf}, \binits{M.}},
\bauthor{\bsnm{Huang}, \binits{X.}},
\bauthor{\bsnm{Booth}, \binits{J.}},
\bauthor{\bsnm{Kramer-Bottiglio}, \binits{R.}},
\bauthor{\bsnm{Balkcom}, \binits{D.}}:
\batitle{Soft lattice modules that behave independently and collectively}.
\bjtitle{IEEE Robotics and Automation Letters}
\bvolume{7}(\bissue{3}),
\bfpage{5942}--\blpage{5949}
(\byear{2022})
\doiurl{10.1109/LRA.2022.3160611}
\end{barticle}
\endbibitem

\bibitem[\protect\citeauthoryear{Xavier et~al.}{2022}]{9785890}
\begin{barticle}
\bauthor{\bsnm{Xavier}, \binits{M.S.}},
\bauthor{\bsnm{Tawk}, \binits{C.D.}},
\bauthor{\bsnm{Zolfagharian}, \binits{A.}},
\bauthor{\bsnm{Pinskier}, \binits{J.}},
\bauthor{\bsnm{Howard}, \binits{D.}},
\bauthor{\bsnm{Young}, \binits{T.}},
\bauthor{\bsnm{Lai}, \binits{J.}},
\bauthor{\bsnm{Harrison}, \binits{S.M.}},
\bauthor{\bsnm{Yong}, \binits{Y.K.}},
\bauthor{\bsnm{Bodaghi}, \binits{M.}},
\bauthor{\bsnm{Fleming}, \binits{A.J.}}:
\batitle{Soft pneumatic actuators: A review of design, fabrication, modeling, sensing, control and applications}.
\bjtitle{IEEE Access}
\bvolume{10},
\bfpage{59442}--\blpage{59485}
(\byear{2022})
\doiurl{10.1109/ACCESS.2022.3179589}
\end{barticle}
\endbibitem

\bibitem[\protect\citeauthoryear{Zhai et~al.}{2023}]{zhai}
\begin{barticle}
\bauthor{\bsnm{Zhai}, \binits{Y.}},
\bauthor{\bsnm{Boer}, \binits{A.D.}},
\bauthor{\bsnm{Yan}, \binits{J.}},
\bauthor{\bsnm{Shih}, \binits{B.}},
\bauthor{\bsnm{Faber}, \binits{M.}},
\bauthor{\bsnm{Speros}, \binits{J.}},
\bauthor{\bsnm{Gupta}, \binits{R.}},
\bauthor{\bsnm{Tolley}, \binits{M.T.}}:
\batitle{Desktop fabrication of monolithic soft robotic devices with embedded fluidic control circuits}.
\bjtitle{Science Robotics}
\bvolume{8}(\bissue{79}),
\bfpage{3792}
(\byear{2023})
\doiurl{10.1126/scirobotics.adg3792}
\end{barticle}
\endbibitem

\bibitem[\protect\citeauthoryear{Tawk and Alici}{2021}]{Tawk2021ARO}
\begin{botherref}
\oauthor{\bsnm{Tawk}, \binits{C.}},
\oauthor{\bsnm{Alici}, \binits{G.}}:
A review of 3d‐printable soft pneumatic actuators and sensors: Research challenges and opportunities.
Advanced Intelligent Systems
\textbf{3}
(2021)
\doiurl{10.1002/aisy.202000223}
\end{botherref}
\endbibitem

\bibitem[\protect\citeauthoryear{Souza et~al.}{2018}]{8658751}
\begin{bchapter}
\bauthor{\bsnm{Souza}, \binits{I.M.L.}},
\bauthor{\bsnm{Andrade}, \binits{W.L.}},
\bauthor{\bsnm{Sampaio}, \binits{L.M.R.}},
\bauthor{\bsnm{Araujo}, \binits{A.L.S.O.}}:
\bctitle{A systematic review on the use of lego robotics in education}.
In: \bbtitle{2018 IEEE Frontiers in Education Conference (FIE)},
pp. \bfpage{1}--\blpage{9}
(\byear{2018}).
\doiurl{10.1109/FIE.2018.8658751}
\end{bchapter}
\endbibitem

\bibitem[\protect\citeauthoryear{Liang et~al.}{2024}]{liang2023decoding}
\begin{barticle}
\bauthor{\bsnm{Liang}, \binits{G.}},
\bauthor{\bsnm{Wu}, \binits{D.}},
\bauthor{\bsnm{Tu}, \binits{Y.}},
\bauthor{\bsnm{Lam}, \binits{T.L.}}:
\batitle{Decoding modular reconfigurable robots: A survey on mechanisms and design}.
\bjtitle{The International Journal of Robotics Research}
(\byear{2024})
\doiurl{10.1177/02783649241283847}
\end{barticle}
\endbibitem

\bibitem[\protect\citeauthoryear{Yim et~al.}{2002}]{Yim2002ModularRobots}
\begin{barticle}
\bauthor{\bsnm{Yim}, \binits{M.}},
\bauthor{\bsnm{Zhang}, \binits{Y.}},
\bauthor{\bsnm{Duff}, \binits{D.}}:
\batitle{Modular robots}.
\bjtitle{IEEE Spectrum}
\bvolume{39}(\bissue{2}),
\bfpage{30}--\blpage{34}
(\byear{2002})
\doiurl{10.1109/6.981854}
\end{barticle}
\endbibitem

\bibitem[\protect\citeauthoryear{Yim et~al.}{2009}]{reconfig_review}
\begin{bbook}
\bauthor{\bsnm{Yim}, \binits{M.}},
\bauthor{\bsnm{White}, \binits{P.}},
\bauthor{\bsnm{Park}, \binits{M.}},
\bauthor{\bsnm{Sastra}, \binits{J.}}:
\bbtitle{Modular Self-Reconfigurable Robots},
pp. \bfpage{5618}--\blpage{5631}.
\bpublisher{Springer},
\blocation{New York, NY}
(\byear{2009}).
\doiurl{10.1007/978-0-387-30440-3_334}
\end{bbook}
\endbibitem

\bibitem[\protect\citeauthoryear{Yim et~al.}{2007}]{4141032}
\begin{barticle}
\bauthor{\bsnm{Yim}, \binits{M.}},
\bauthor{\bsnm{Shen}, \binits{W.-m.}},
\bauthor{\bsnm{Salemi}, \binits{B.}},
\bauthor{\bsnm{Rus}, \binits{D.}},
\bauthor{\bsnm{Moll}, \binits{M.}},
\bauthor{\bsnm{Lipson}, \binits{H.}},
\bauthor{\bsnm{Klavins}, \binits{E.}},
\bauthor{\bsnm{Chirikjian}, \binits{G.S.}}:
\batitle{Modular self-reconfigurable robot systems [grand challenges of robotics]}.
\bjtitle{IEEE Robotics Automation Magazine}
\bvolume{14}(\bissue{1}),
\bfpage{43}--\blpage{52}
(\byear{2007})
\doiurl{10.1109/MRA.2007.339623}
\end{barticle}
\endbibitem

\bibitem[\protect\citeauthoryear{Liu et~al.}{2019}]{8769941}
\begin{barticle}
\bauthor{\bsnm{Liu}, \binits{C.}},
\bauthor{\bsnm{Whitzer}, \binits{M.}},
\bauthor{\bsnm{Yim}, \binits{M.}}:
\batitle{A distributed reconfiguration planning algorithm for modular robots}.
\bjtitle{IEEE Robotics and Automation Letters}
\bvolume{4}(\bissue{4}),
\bfpage{4231}--\blpage{4238}
(\byear{2019})
\doiurl{10.1109/LRA.2019.2930432}
\end{barticle}
\endbibitem

\bibitem[\protect\citeauthoryear{Wei et~al.}{2010}]{wei2010Sambot}
\begin{bchapter}
\bauthor{\bsnm{Wei}, \binits{H.}},
\bauthor{\bsnm{Cai}, \binits{Y.}},
\bauthor{\bsnm{Li}, \binits{H.}},
\bauthor{\bsnm{Li}, \binits{D.}},
\bauthor{\bsnm{Wang}, \binits{T.}}:
\bctitle{Sambot: A self-assembly modular robot for swarm robot}.
In: \bbtitle{2010 IEEE International Conference on Robotics and Automation},
pp. \bfpage{66}--\blpage{71}
(\byear{2010}).
\doiurl{10.1109/ROBOT.2010.5509214}
\end{bchapter}
\endbibitem

\bibitem[\protect\citeauthoryear{Ozkan-Aydin and Goldman}{2021}]{multi_leg_swarm}
\begin{barticle}
\bauthor{\bsnm{Ozkan-Aydin}, \binits{Y.}},
\bauthor{\bsnm{Goldman}, \binits{D.I.}}:
\batitle{Self-reconfigurable multilegged robot swarms collectively accomplish challenging terradynamic tasks}.
\bjtitle{Science Robotics}
\bvolume{6}(\bissue{56}),
\bfpage{1628}
(\byear{2021})
\doiurl{10.1126/scirobotics.abf1628}
\end{barticle}
\endbibitem

\bibitem[\protect\citeauthoryear{Freeman et~al.}{2023}]{freeman2023topology}
\begin{barticle}
\bauthor{\bsnm{Freeman}, \binits{C.}},
\bauthor{\bsnm{Maynard}, \binits{M.}},
\bauthor{\bsnm{Vikas}, \binits{V.}}:
\batitle{Topology and morphology design of spherically reconfigurable homogeneous modular soft robots}.
\bjtitle{Soft Robotics}
\bvolume{10}(\bissue{1}),
\bfpage{52}--\blpage{65}
(\byear{2023})
\doiurl{10.1089/soro.2021.0125}
\end{barticle}
\endbibitem

\bibitem[\protect\citeauthoryear{Li and Rus}{2019}]{Li2019JelloCubeAC}
\begin{barticle}
\bauthor{\bsnm{Li}, \binits{S.}},
\bauthor{\bsnm{Rus}, \binits{D.}}:
\batitle{Jellocube: A continuously jumping robot with soft body}.
\bjtitle{IEEE/ASME Transactions on Mechatronics}
\bvolume{24}(\bissue{2}),
\bfpage{447}--\blpage{458}
(\byear{2019})
\doiurl{10.1109/TMECH.2019.2899606}
\end{barticle}
\endbibitem

\bibitem[\protect\citeauthoryear{Zou et~al.}{2018}]{zou2018reconfigurable}
\begin{barticle}
\bauthor{\bsnm{Zou}, \binits{J.}},
\bauthor{\bsnm{Lin}, \binits{Y.}},
\bauthor{\bsnm{Ji}, \binits{C.}},
\bauthor{\bsnm{Yang}, \binits{H.}}:
\batitle{A reconfigurable omnidirectional soft robot based on caterpillar locomotion}.
\bjtitle{Soft Robotics}
\bvolume{5}(\bissue{2}),
\bfpage{164}--\blpage{174}
(\byear{2018})
\doiurl{10.1089/soro.2017.0008}
\end{barticle}
\endbibitem

\bibitem[\protect\citeauthoryear{Suzuki et~al.}{2019}]{ShapeBotsSuzukiEtAl}
\begin{bchapter}
\bauthor{\bsnm{Suzuki}, \binits{R.}},
\bauthor{\bsnm{Zheng}, \binits{C.}},
\bauthor{\bsnm{Kakehi}, \binits{Y.}},
\bauthor{\bsnm{Yeh}, \binits{T.}},
\bauthor{\bsnm{Do}, \binits{E.Y.-L.}},
\bauthor{\bsnm{Gross}, \binits{M.D.}},
\bauthor{\bsnm{Leithinger}, \binits{D.}}:
\bctitle{Shapebots: Shape-changing swarm robots}.
In: \bbtitle{Proceedings of the 32nd Annual ACM Symposium on User Interface Software and Technology}.
\bsertitle{UIST '19},
pp. \bfpage{493}--\blpage{505}.
\bpublisher{Association for Computing Machinery},
\blocation{New York, NY, USA}
(\byear{2019}).
\doiurl{10.1145/3332165.3347911}
\end{bchapter}
\endbibitem

\bibitem[\protect\citeauthoryear{Ceron et~al.}{2021}]{Foambots}
\begin{barticle}
\bauthor{\bsnm{Ceron}, \binits{S.}},
\bauthor{\bsnm{Kimmel}, \binits{M.A.}},
\bauthor{\bsnm{Nilles}, \binits{A.}},
\bauthor{\bsnm{Petersen}, \binits{K.}}:
\batitle{Soft robotic oscillators with strain-based coordination}.
\bjtitle{IEEE Robotics and Automation Letters}
\bvolume{6}(\bissue{4}),
\bfpage{7557}--\blpage{7563}
(\byear{2021})
\doiurl{10.1109/LRA.2021.3100599}
\end{barticle}
\endbibitem

\bibitem[\protect\citeauthoryear{Malley et~al.}{2020}]{eciton}
\begin{bchapter}
\bauthor{\bsnm{Malley}, \binits{M.}},
\bauthor{\bsnm{Haghighat}, \binits{B.}},
\bauthor{\bsnm{Houel}, \binits{L.}},
\bauthor{\bsnm{Nagpal}, \binits{R.}}:
\bctitle{Eciton robotica: Design and algorithms for an adaptive self-assembling soft robot collective}.
In: \bbtitle{2020 IEEE International Conference on Robotics and Automation (ICRA)},
pp. \bfpage{4565}--\blpage{4571}
(\byear{2020}).
\doiurl{10.1109/ICRA40945.2020.9196565}
\end{bchapter}
\endbibitem

\bibitem[\protect\citeauthoryear{Li et~al.}{2022}]{li2022scaling}
\begin{barticle}
\bauthor{\bsnm{Li}, \binits{S.}},
\bauthor{\bsnm{Awale}, \binits{S.A.}},
\bauthor{\bsnm{Bacher}, \binits{K.E.}},
\bauthor{\bsnm{Buchner}, \binits{T.J.}},
\bauthor{\bsnm{Della~Santina}, \binits{C.}},
\bauthor{\bsnm{Wood}, \binits{R.J.}},
\bauthor{\bsnm{Rus}, \binits{D.}}:
\batitle{Scaling up soft robotics: A meter-scale, modular, and reconfigurable soft robotic system}.
\bjtitle{Soft Robotics}
\bvolume{9}(\bissue{2}),
\bfpage{324}--\blpage{336}
(\byear{2022})
\doiurl{10.1089/soro.2020.0123}
\end{barticle}
\endbibitem

\bibitem[\protect\citeauthoryear{Zhang et~al.}{2020}]{ModularSR}
\begin{botherref}
\oauthor{\bsnm{Zhang}, \binits{C.}},
\oauthor{\bsnm{Zhu}, \binits{P.}},
\oauthor{\bsnm{Lin}, \binits{Y.}},
\oauthor{\bsnm{Jiao}, \binits{Z.}},
\oauthor{\bsnm{Zou}, \binits{J.}}:
Modular soft robotics: Modular units, connection mechanisms, and applications.
Advanced Intelligent Systems
\textbf{2}
(2020)
\doiurl{10.1002/aisy.201900166}
\end{botherref}
\endbibitem

\bibitem[\protect\citeauthoryear{Lee et~al.}{2024}]{2024_Lee}
\begin{botherref}
\oauthor{\bsnm{Lee}, \binits{H.C.}},
\oauthor{\bsnm{Elder}, \binits{N.}},
\oauthor{\bsnm{Leal}, \binits{M.}},
\oauthor{\bsnm{Stantial}, \binits{S.}},
\oauthor{\bsnm{Martinez}, \binits{E.V.}},
\oauthor{\bsnm{Jos}, \binits{S.}},
\oauthor{\bsnm{Cho}, \binits{H.}},
\oauthor{\bsnm{Russo}, \binits{S.}}:
A fabrication strategy for millimeter-scale, self-sensing soft-rigid hybrid robots.
Nature Communications
\textbf{15}(1)
(2024)
\doiurl{10.1038/s41467-024-51137-8}
\end{botherref}
\endbibitem

\bibitem[\protect\citeauthoryear{Usevitch et~al.}{2020}]{untethered_isoperimetric}
\begin{barticle}
\bauthor{\bsnm{Usevitch}, \binits{N.S.}},
\bauthor{\bsnm{Hammond}, \binits{Z.M.}},
\bauthor{\bsnm{Schwager}, \binits{M.}},
\bauthor{\bsnm{Okamura}, \binits{A.M.}},
\bauthor{\bsnm{Hawkes}, \binits{E.W.}},
\bauthor{\bsnm{Follmer}, \binits{S.}}:
\batitle{An untethered isoperimetric soft robot}.
\bjtitle{Science Robotics}
\bvolume{5}(\bissue{40}),
\bfpage{0492}
(\byear{2020})
\doiurl{10.1126/scirobotics.aaz0492}
\end{barticle}
\endbibitem

\bibitem[\protect\citeauthoryear{Schmitt et~al.}{2018}]{schmitt2018soft}
\begin{botherref}
\oauthor{\bsnm{Schmitt}, \binits{F.}},
\oauthor{\bsnm{Piccin}, \binits{O.}},
\oauthor{\bsnm{Barbé}, \binits{L.}},
\oauthor{\bsnm{Bayle}, \binits{B.}}:
Soft robots manufacturing: A review.
Frontiers in Robotics and AI
\textbf{5}
(2018)
\doiurl{10.3389/frobt.2018.00084}
\end{botherref}
\endbibitem

\bibitem[\protect\citeauthoryear{Wehner et~al.}{2016}]{wehner2016integrated}
\begin{barticle}
\bauthor{\bsnm{Wehner}, \binits{M.}},
\bauthor{\bsnm{Truby}, \binits{R.}},
\bauthor{\bsnm{Fitzgerald}, \binits{D.}},
\bauthor{\bsnm{Mosadegh}, \binits{B.}},
\bauthor{\bsnm{Whitesides}, \binits{G.}},
\bauthor{\bsnm{Lewis}, \binits{J.}},
\bauthor{\bsnm{Wood}, \binits{R.}}:
\batitle{An integrated design and fabrication strategy for entirely soft, autonomous robots}.
\bjtitle{Nature}
\bvolume{536},
\bfpage{451}--\blpage{455}
(\byear{2016})
\doiurl{10.1038/nature19100}
\end{barticle}
\endbibitem

\end{thebibliography}

\backmatter

\newpage




\textbf{Acknowledgments:} We thank the creators of \textit{SeaAnimal4k} (Star fish 4k Amazing Starfish in Undersea Ultra Hd, \url{https://www.youtube.com/watch?v=MWlKfaxROd0}), \textit{keepondiving3140} (Black brittle star Small Giftun Red Sea Egypt, \url{https://www.youtube.com/watch?v=cX9c43sF3vo}), and \textit{SnakeDiscovery} (How Snakes Move! (They don't just slither!), \url{https://www.youtube.com/watch?v=7-AKPFiIEEw&t=30s}) for their contribution to providing high-quality references for the locomotion of different creatures: provided videos of starfish, brittle star, and snake locomotion.

\textbf{Author contributions:} Conceptualization: LZ, DB, MC, CS, YJ; Methodology: LZ, DB, MC, CS, YJ; Investigation: LZ, DB, MC, CS, YJ; Visualization: LZ, YJ; Experiments Design and Implementation: LZ, DB, MC, CS, YJ; System Design and Implementation: LZ, DB, MC, CS, YJ; Algorithm Design and Implementation: LZ, DB, MC, YJ, CS; Project administration: DB, LZ; Supervision: DB, MC; Writing: original draft: LZ; Writing: review \& editing: LZ, DB, MC, CS, YJ. 

\textbf{Competing interests:} The authors declare they have no competing interests. 


\clearpage 
\section*{Supplementary materials}

\textbf{This PDF file includes:}

\noindent Supplementary methods

\noindent Figures S1 to S3

\noindent Algorithm 1


\noindent \textbf{Other supplementary material}

\noindent Movie S1







\section*{Supplementary methods}

\underline{String length calculation}


The string length \( L \) between two ribs is parameterized by the bending angle \( \alpha \) and the threading locations on both ribs. The distances from the threading points on ribs 1 and 2 to the rib's endpoint (which is closest to the center \((x_0, y_0)\)) are defined as \( l_1 \) and \( l_2 \), respectively. The corresponding threading positions are represented by the one-dimensional scalars \( \lambda_1 \) and \( \lambda_2 \), which are calculated by subtracting half the rib's length from \( l_1 \) and \( l_2 \). Specifically, \( \lambda_1 = l_1 - L_{\text{rib}}/2 \) and \( \lambda_2 = l_2 - L_{\text{rib}}/2 \), an example is shown in Figure S1.

\textbf{Case 1}: Straight line (\( \alpha \approx 0 \)). When \( \alpha \) is small, the beam connecting the ribs remains nearly straight. In this case, the string length \( L \) is the Euclidean distance between the threading points \( S_1 \) and \( S_2 \), calculated as:
\begin{align}
S_1 & = \left( x_1 + \lambda_1  \cos{\theta_1}, \, y_1 + \lambda_1  \sin{\theta_1} \right),\\
S_2 & =  \left( x_1 + \lambda_2  \cos{(\theta_1)} - s \sin{(\theta_1)},  y_1 + \lambda_2  \sin{(\theta_1)} + s \cos{(\theta_1)} \right).
\end{align}

\textbf{Case 2}: Curved arc (\( \alpha \neq 0 \)). For non-zero \( \alpha \), the ribs form a circular arc. The radius \( R \) of the arc is determined by the relation: $R = s/|\alpha|$. The center of the arc is located at:
$x_0 = x_1 - R  \cos{\theta_1}$ and $y_0 = y_1 - R  \sin{\theta_1}$. The endpoint of the arc \( (x_2, y_2) \) is positioned at an angle \( \theta_1 + \alpha \), and is calculated as: $x_2 = x_0 + R \cos(\theta_1 + \alpha)$ and $y_2 = y_0 + R \sin(\theta_1 + \alpha).$ 
The threading points \( S_1 \) and \( S_2 \) on the arc are determined by:
\begin{align}
S_1 = ( x_1 + \lambda_1  \frac{x_1 - x_0}{R}, \, y_1 + \lambda_1  \frac{y_1 - y_0}{R} ),\\
S_2 = ( x_2 + \lambda_2  \frac{x_2 - x_0}{R}, \, y_2 + \lambda_2  \frac{y_2 - y_0}{R} ).
\end{align}

Please note that there are two cases because when \( \alpha \approx 0 \), the values \( x_1 - x_0 \), \( x_2 - x_0 \), and \( R \) will all diverge to infinity, causing numerical instability due to floating-point precision limits. To address this, a blending method is used to achieve greater numerical stability. When the radius of the arc becomes very large, the arc behaves more like a straight line, and the trigonometric calculations based on the arc center lose accuracy. Thus, for those cases, we switched to a linear approximation based on the arc's starting point and length, where the coordinates of \( x_2 \) and \( y_2 \) are given by \( x_2 = x_1 + s \cos(\theta_1 + \alpha + \frac{\pi}{2}) \) and \( y_2 = y_1 + s \sin(\theta_1 + \alpha + \frac{\pi}{2}) \). This method ensures that even for large radii, the endpoints of the arc are computed efficiently and without the risk of crossing or overlap.
For both cases, the string length \( L \) is given by:
\begin{align}\label{L_case1}
L = ||S_1-S_2||.
\end{align}

\begin{algorithm}\label{algorithm1}
\caption{Quasi-static model.}
\begin{algorithmic} 
\State \textbf{Initialization:} $\epsilon = 1 \times 10^{-6}$
\While{$|L_{i,k} - L_{i,k-1}| \leq \epsilon$}
    \State Optimized $L_{i}^*$ from:
    \State $\begin{cases}
    {\text{min}} \sum \alpha_i^2 = \textit{Length2Angle}(L_i, T_i)
    \\
    \text{s.t. } \sum L_{i} = L_{total} 
    \end{cases}$
    \State $k \gets k+1$
\EndWhile
\Function{$\alpha$ = \textit{Length2Angle}($L_i, T_i$)}{}
    \State $\alpha \gets \emptyset$
    \For{$i = 0 : n$}
        \State Optimized $\alpha_i$ from:
        \State $\begin{cases}
        \underset{\alpha_i}{\text{min}} \quad ||L(\alpha_i) - L_i|| \\
        \text{s.t. } -\pi < \alpha_i \leq \pi 
        \end{cases}$
        \State $\alpha \gets \alpha \cup \{\alpha_i\}$
    \EndFor
\EndFunction
\end{algorithmic}
\end{algorithm}

\noindent\textbf{Supplementary Figures}
\vspace*{2mm}

\rule{0pt}{5mm} 
\noindent\begin{minipage}{\linewidth}
\centering
\includegraphics[trim=0cm 6.5cm 0cm 6cm, clip=true, width=\linewidth]{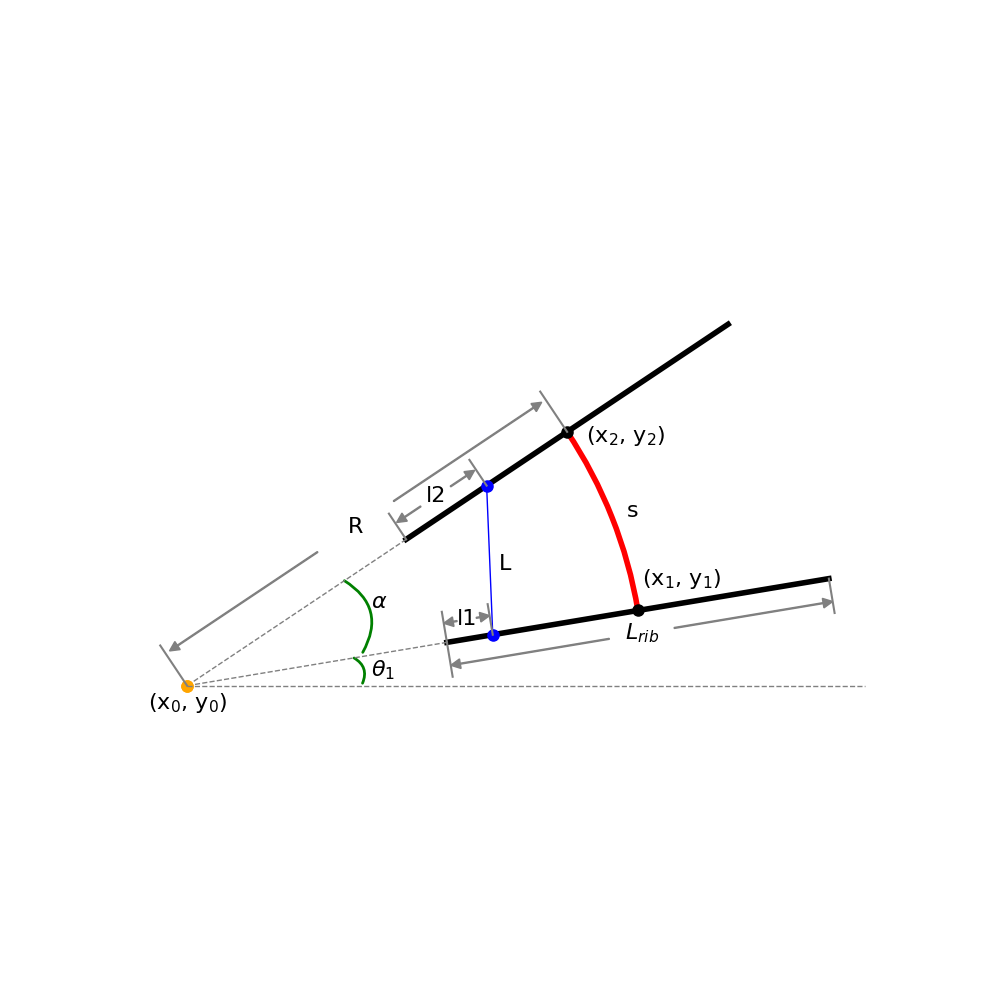}
\rule{0pt}{5mm} 
\parbox{\linewidth}{\textbf{Figure S1: Example showing the deformation behavior of a segment of the skeleton.} The thick black lines represent the ribs of the passive skeleton, with points $(x_1, y_1)$ and $(x_2, y_2)$ located at the center of the bottom and top ribs, respectively. The length of the rib is $L_{rib}$. The thick red line signifies the curved segments between the ribs along the skeleton's longitudinal direction, having an arc length $s$, which represents the spine of the skeleton. The radius of the spine-bending curve is denoted by $R$.
The thin blue line indicates the string connecting the threading points, with length $L$. The angle $\theta_1$ describes the orientation of the bottom rib relative to the horizontal, and $\alpha$ is the angle between the two ribs. 
The distances from the threading points on the bottom rib and top rib to each rib's endpoint, which is closest to the center \((x_0, y_0)\), are defined as \( l_1 \) and \( l_2 \), respectively.
}

\end{minipage}

\newpage
\rule{0pt}{5mm} 
\noindent\begin{minipage}{\linewidth}
\centering
\includegraphics[trim=0cm 0cm 0cm 0cm, clip=true, width=\linewidth]{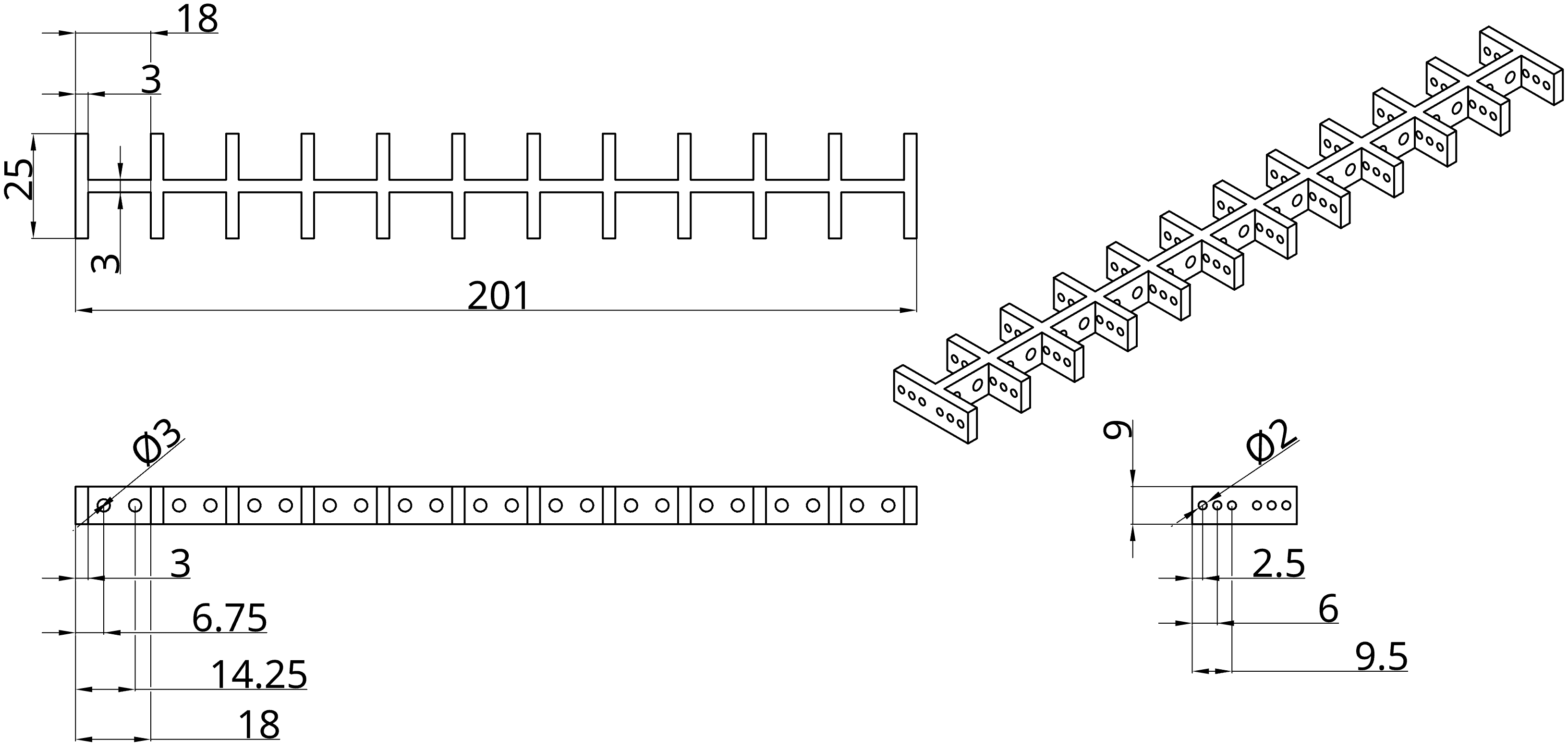}
\rule{0pt}{5mm} 
\parbox{\linewidth}{{\textbf{Figure S2: Specifications of the 3D-printed skeleton.} The front, side, top, and axonometric drawings with dimensions are displayed, showing a total length of 201 mm, width of 25 mm, and thickness of 9 mm. Each vertical rib features six threading holes, with an additional two holes located between the ribs. This arrangement of holes provides design flexibility, allowing for the use of a single string to thread through the holes in various configurations. This enables the creation of diverse shapes with SoftSnap modules. By designing customized connectors, these modules can be assembled into innovative, mass-produced soft robots.}}
\end{minipage}

\newpage
\rule{0pt}{5mm} 
\noindent\begin{minipage}{\linewidth}
\centering
\includegraphics[trim=0cm 0cm 0cm 0cm, clip=true, width=\linewidth]{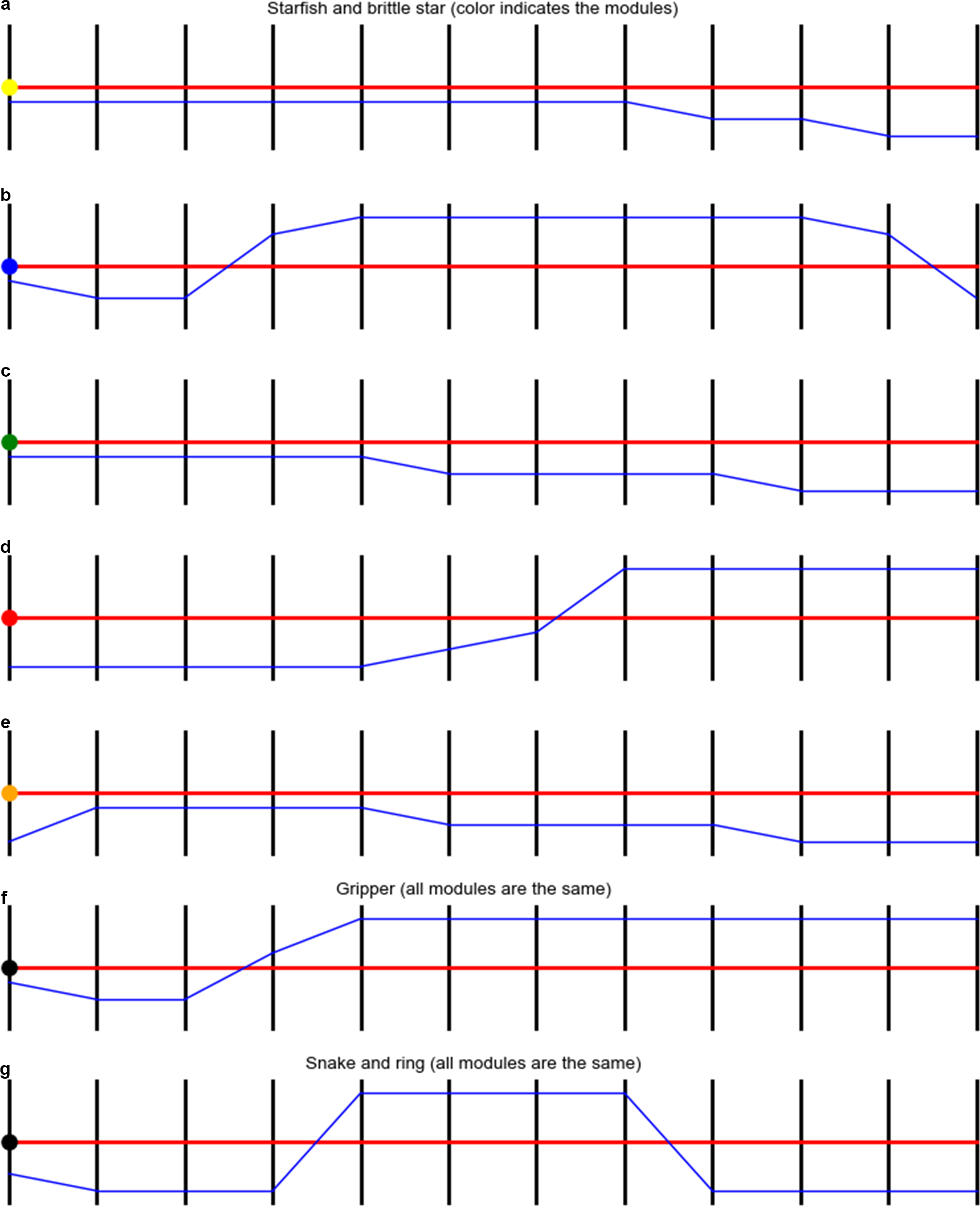}
\rule{0pt}{5mm} 
\parbox{\linewidth}{{\textbf{Figure S3: Threading details for all the robots in this paper}. (a)-(e) the five SoftSnap modules in yellow, blue, green, red, and orange used in the Starfish and Brittle Star robots, (f) the four SoftSnap modules in the Gripper robot, and (g) the four SoftSnap modules in the Snake and Ring robots. The black and red lines depict the passive skeleton, and the blue lines are strings running through holes from the left to the right end of the skeleton.}}
\end{minipage}


\newpage
\rule{0pt}{5mm} 
\noindent\begin{minipage}{\linewidth}
\centering
\includegraphics[width=\linewidth]{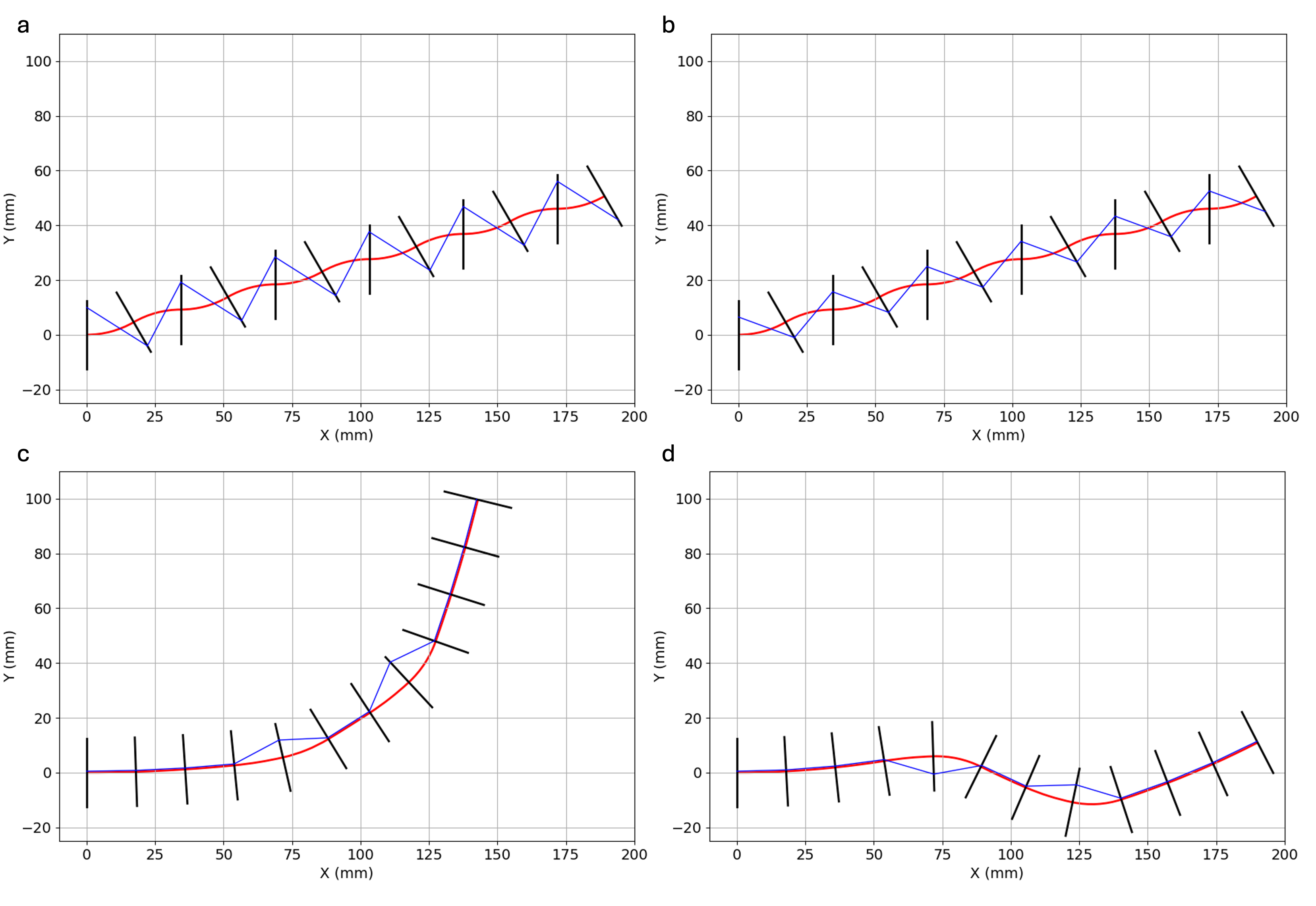}
\rule{0pt}{5mm} 
\parbox{\linewidth}{{\textbf{Figure S4: Calculated threading pattern and string lengths given the desired angles}. (a) Desired angles: [30$^\circ$, -30$^\circ$, 30$^\circ$, -30$^\circ$, 30$^\circ$, -30$^\circ$, 30$^\circ$, -30$^\circ$, 30$^\circ$, -30$^\circ$, 30$^\circ$], obtained angles: [30$^\circ$, -30$^\circ$, 30$^\circ$, -30$^\circ$, 30$^\circ$, -30$^\circ$, 30$^\circ$, -30$^\circ$, 30$^\circ$, -30$^\circ$, 30$^\circ$], threading patterns: [-10, 10, -10, 10, -10, 10, -10, 10, -10, 10, -10, 10] mm, string lengths: 270 mm. (b) Desired angles: [30$^\circ$, -30$^\circ$, 30$^\circ$, -30$^\circ$, 30$^\circ$, -30$^\circ$, 30$^\circ$, -30$^\circ$, 30$^\circ$, -30$^\circ$, 30$^\circ$], obtained angles: [30$^\circ$, -30$^\circ$, 30$^\circ$, -30$^\circ$, 30$^\circ$, -30$^\circ$, 30$^\circ$, -30$^\circ$, 30$^\circ$, -30$^\circ$, 30$^\circ$], threading patterns: [-6.5, 6.5, -6.5, 6.5, -6.5, 6.5, -6.5, 6.5, -6.5, 6.5, -6.5, 6.5] mm, string lengths: 224 mm. (c) Desired angles: [0$^\circ$, 0$^\circ$, 0$^\circ$, 10$^\circ$, 15$^\circ$, 0$^\circ$, 10$^\circ$, 30$^\circ$, 0$^\circ$, 0$^\circ$, 0$^\circ$], obtained angles: [1.9$^\circ$, 1.9$^\circ$, 1.9$^\circ$, 7.2$^\circ$, 18.6$^\circ$, 1.9$^\circ$, 9.7$^\circ$, 27.7$^\circ$, 1.9$^\circ$, 1.9$^\circ$, 1.9$^\circ$], threading patterns: [-0.5, -0.5, -0.5, -0.5, -6.5, -0.5, -0.5, -10, -0.5, -0.5, -0.5, -0.5] mm, string lengths: 199 mm. (d) Desired angles: [0$^\circ$, 0$^\circ$, 0$^\circ$, -10$^\circ$, -30$^\circ$, 0$^\circ$, 10$^\circ$, 30$^\circ$, 0$^\circ$, 0$^\circ$, 0$^\circ$], obtained angles: [3$^\circ$, 3$^\circ$, 3$^\circ$$^\circ$, -7.2$^\circ$, -28.1$^\circ$, 3$^\circ$, 11.6$^\circ$, 29.9$^\circ$, 3$^\circ$, 3$^\circ$, 3$^\circ$], threading patterns: [-0.5, -0.5, -0.5, -0.5, 6.5, -0.5, -0.5, -6.5, -0.5, -0.5, -0.5, -0.5] mm, string lengths: 196 mm.}}
\end{minipage}

\end{document}